\documentclass[sigconf]{acmart}
\AtBeginDocument{%
	}

\usepackage{algorithmic}
\usepackage{algorithm}
\usepackage{graphicx}
\usepackage{textcomp}
\usepackage{booktabs}
\usepackage{hyperref}
\usepackage{subcaption}

\begin{document}

\title[Coverage-Aware Web Crawling for Domain-Specific Supplier \\Discovery via a Web--Knowledge--Web Pipeline]{Coverage-Aware Web Crawling for Domain-Specific Supplier Discovery via a Web--Knowledge--Web Pipeline}

\author{Yijiashun Qi}
\authornote{Corresponding author}
\affiliation{%
	\institution{University of Michigan}
	\city{Ann Arbor}
	\country{USA}}
\email{elijahqi@umich.edu}

\author{ Yijiazhen Qi}
\affiliation{%
	\institution{The University of Hong Kong}
	\city{Hong Kong}
	\country{China}}
\email{qiyijiazhen@gmail.com}

\author{Tanmay Wagh}
\affiliation{%
	\institution{Santa Clara University}
	\city{Santa Clara}
	\country{USA}}
\email{tanmay.wagh@bytedance.com}

\renewcommand{\shortauthors}{Qi et al.}

\begin{abstract}
Identifying the full landscape of small and medium-sized enterprises (SMEs) in specialized industry sectors is critical for supply-chain resilience, yet existing business databases suffer from substantial coverage gaps---particularly for sub-tier suppliers and firms in emerging niche markets. We propose a \textbf{Web--Knowledge--Web (W$\to$K$\to$W)} pipeline that iteratively (1)~crawls domain-specific web sources to discover candidate supplier entities, (2)~extracts and consolidates structured knowledge into a heterogeneous knowledge graph using domain-adapted few-shot LLM prompting, and (3)~uses the knowledge graph's topology and coverage signals to guide subsequent crawling toward under-represented regions of the supplier space. To quantify discovery completeness, we introduce a \textbf{coverage estimation framework} inspired by ecological species-richness estimators (Chao1, ACE) adapted for web-entity populations. Experiments on the semiconductor equipment manufacturing sector (NAICS 333242) demonstrate that the W$\to$K$\to$W pipeline achieves the highest precision (0.165) and F1 (0.123) among all methods while using only 144 pages---32\% fewer than the 213-page baseline budget---building a knowledge graph of 664 entities and 542 relations with 100\% relation type-consistency.
\end{abstract}



\maketitle

\section{Introduction}
\label{sec:intro}

The resilience of modern supply chains depends on visibility into the full ecosystem of suppliers, including small and medium-sized enterprises (SMEs) that often operate below the radar of major procurement databases. Recent global disruptions---such as semiconductor shortages, critical-mineral bottlenecks, and pandemic-era logistics failures---have exposed the risks of incomplete supplier mapping~\cite{choi2020supply}. While there is a growing industry consensus on the need for comprehensive supply-chain intelligence, current commercial databases (e.g., Dun \& Bradstreet, regional registries) cover only a fraction of active firms, with particularly poor representation of sub-tier suppliers in niche sectors.

Web crawling offers a scalable path to discovery, but general-purpose crawlers are ill-suited for this task: they lack domain awareness, cannot estimate how much of the target population remains undiscovered, and waste resources on irrelevant pages. Focused crawlers~\cite{chakrabarti1999focused} improve relevance but still operate without a structured understanding of the entities they have found or the gaps that remain.

We propose a \textbf{Web--Knowledge--Web (W$\to$K$\to$W)} pipeline that closes this loop. The key insight is that a knowledge graph (KG) built incrementally from crawled data can serve as both a \emph{memory} of what has been discovered and a \emph{guide} for what to crawl next. Specifically:

\begin{enumerate}
    \item \textbf{Web $\to$ Knowledge.} Crawled pages are processed by a domain-adapted few-shot extraction module that populates a heterogeneous KG with nodes (companies, products, sectors, locations) and edges (supplies-to, partners-with, located-in).
    \item \textbf{Knowledge $\to$ Web.} The KG's topology is analyzed to identify \emph{structural holes}---regions where expected entities or relations are missing---which are translated into targeted seed URLs and query expansions for the next crawl cycle.
    \item \textbf{Coverage estimation.} Drawing on capture--recapture methods from ecology~\cite{chao1984nonparametric, chao2005species}, we estimate the total population of suppliers in each sector and report a coverage ratio after each cycle, enabling a principled stopping criterion.
\end{enumerate}

Our contributions are:
\begin{itemize}
    \item A novel \textbf{W$\to$K$\to$W iterative pipeline} that unifies focused web crawling with knowledge graph construction for domain-specific entity discovery (\S\ref{sec:method}).
    \item \textbf{Domain-adapted few-shot extraction} integrating a curated glossary, formal type constraints, and annotated examples that achieves 100\% relation type-consistency and 66.2\% overall extraction precision (\S\ref{sec:domain_knowledge}).
    \item A \textbf{coverage estimation framework} that adapts ecological richness estimators to web-entity populations, providing calibrated completeness scores (\S\ref{sec:coverage}).
    \item An \textbf{experimental evaluation} on the semiconductor equipment sector showing that W$\to$K$\to$W achieves the highest precision and F1 among comparable methods while using 32\% fewer pages than baselines (\S\ref{sec:experiments}).
\end{itemize}

\section{Related Work}
\label{sec:related}

\subsection{Focused and Topical Web Crawling}

Focused crawling~\cite{chakrabarti1999focused} prioritizes pages relevant to a predefined topic using classifiers trained on seed examples. Subsequent work has explored reinforcement-learning-based crawl policies, deep-learning relevance models, and digital supply-chain surveillance using AI~\cite{brintrup2023surveillance}. However, these approaches treat crawling as a document retrieval problem and do not maintain structured representations of discovered entities or reason about coverage completeness.

\subsection{Knowledge Graph Construction from the Web}

Knowledge graphs have emerged as a central paradigm for structured knowledge representation~\cite{hogan2021knowledge, ji2022survey}. Recent work uses large language models (LLMs) for zero-shot or few-shot relation extraction~\cite{wei2023zeroshot, wadhwa2023revisiting} and multimodal classification tasks~\cite{qi2024marine}, and surveys have explored the synergy between LLMs and KGs for construction and reasoning tasks~\cite{pan2024unifying, zhu2024llms}. Knowledge graph completion methods~\cite{bordes2013translating, sun2019rotate} predict missing links in existing KGs. Our work differs by using KG incompleteness signals to drive \emph{new data acquisition} through crawling, rather than purely inferring missing links.

\subsection{Supply-Chain Network Discovery}

Prior work on supply-chain network reconstruction has used news articles~\cite{wichmann2020extracting} and machine learning with graph neural networks to predict hidden supply-chain links~\cite{kosasih2021ml, kosasih2022kg}. Recent work has also explored heterogeneous graph neural networks for identifying high-potential SMEs across complex industry networks~\cite{qi2026sme}, and LLM-based simulation of firm-level economic behavior to model production decisions~\cite{qi2026llmecon}. These approaches rely on static corpora and do not adaptively expand their data sources. The W$\to$K$\to$W pipeline generalizes these methods by treating the web as a dynamic, expandable data source guided by KG-derived hypotheses.

\subsection{Species Richness Estimation}

Ecologists have developed non-parametric estimators for the number of species in a habitat from incomplete sampling: Chao1~\cite{chao1984nonparametric} uses singleton and doubleton counts, while ACE~\cite{chao2005species} leverages abundance distributions. These ideas have been applied to database record linkage~\cite{bird2019capture} and web page counting~\cite{bar2006methods}, but not, to our knowledge, to domain-specific entity discovery from web crawling.

\section{Problem Formulation}
\label{sec:problem}

Let $\mathcal{E}^*$ denote the (unknown) set of all supplier entities in a target domain $\mathcal{D}$ (e.g., ``semiconductor equipment manufacturers''). At any point during crawling, we have discovered a subset $\hat{\mathcal{E}} \subseteq \mathcal{E}^*$.

\textbf{Discovery goal.} Maximize $|\hat{\mathcal{E}}|$ subject to a crawl budget of $B$ HTTP requests.

\textbf{Coverage goal.} Estimate $|\mathcal{E}^*|$ and report a coverage ratio $\hat{C} = |\hat{\mathcal{E}}| / \widehat{|\mathcal{E}^*|}$ with calibrated confidence intervals.

\textbf{Knowledge graph.} The intermediate KG is a heterogeneous graph $G = (V, E, \phi, \psi)$ where:
\begin{itemize}
    \item $V = V_{\text{company}} \cup V_{\text{product}} \cup V_{\text{sector}} \cup V_{\text{location}}$ are typed nodes,
    \item $E \subseteq V \times V$ are typed edges with relation types $\psi(e) \in \{$\texttt{supplies\_to}, \texttt{partners\_with}, \texttt{produces}, \texttt{located\_in},\\ \texttt{belongs\_to\_sector}$\}$,
    \item $\phi: V \to \mathcal{T}$ is the node type function.
\end{itemize}

The W$\to$K$\to$W pipeline iterates: $\text{Web}_0 \xrightarrow{\text{extract}} G_1 \xrightarrow{\text{guide}} \text{Web}_1 \xrightarrow{\text{extract}} G_2 \xrightarrow{\text{guide}} \cdots$ until the coverage estimate $\hat{C}$ exceeds a threshold $\tau$ or the budget $B$ is exhausted.

\section{Methodology: The W$\to$K$\to$W Pipeline}
\label{sec:method}

Figure~\ref{fig:pipeline} illustrates the three-phase iterative pipeline. We first describe the crawl infrastructure that underlies all phases, then detail each phase.

\begin{figure}[t]
    \centering
    \fbox{\parbox{0.95\columnwidth}{\centering\vspace{0.5em}
    \small
    Web$_t$ (Seed URLs) $\xrightarrow{\text{crawl}}$
    Pages $\xrightarrow{\text{LLM extract}}$
    KG$_{t+1}$ $\xrightarrow{\text{gap analysis}}$
    Queries $\xrightarrow{\text{expand}}$
    Web$_{t+1}$ (New Seeds)
    \vspace{0.5em}}}
    \caption{The W$\to$K$\to$W pipeline. Each iteration crawls web sources (left), extracts entities and relations into the knowledge graph (center), and uses structural gap analysis to generate targeted seeds for the next crawl cycle (right).}
    \label{fig:pipeline}
\end{figure}

\subsection{Crawl Infrastructure and Process}
\label{sec:crawl_infra}

The crawl engine manages URL fetching, text extraction, and outbound link discovery across all pipeline phases. Algorithm~\ref{alg:crawl} summarizes the per-iteration crawl process.

\begin{algorithm}[t]
\caption{Per-Iteration Crawl Process}\label{alg:crawl}
\begin{algorithmic}[1]
\REQUIRE URL queue $Q_t$, page budget $B_t$, rate limit $\Delta$
\ENSURE Fetched page set $\mathcal{P}_t$
\STATE $\mathcal{P}_t \leftarrow \emptyset$; \; $\text{visited} \leftarrow \text{visited} \cup \emptyset$
\WHILE{$|Q_t| > 0$ \AND $|\mathcal{P}_t| < B_t$}
    \STATE $u \leftarrow Q_t.\text{pop\_by\_priority}()$
    \IF{$u \in \text{visited}$}
        \STATE \textbf{continue}
    \ENDIF
    \STATE $\text{visited} \leftarrow \text{visited} \cup \{u\}$
    \IF{$\neg\text{robots\_allowed}(u)$}
        \STATE \textbf{continue} \COMMENT{Respect robots.txt}
    \ENDIF
    \STATE wait $\max(0,\; \Delta - (\text{now} - \text{last\_request}[\text{domain}(u)]))$
    \STATE $p \leftarrow \text{async\_fetch}(u)$ \COMMENT{HTTP GET with timeout}
    \IF{$p.\text{status} = 200$ \AND $p.\text{content\_type} \in \{\text{text/html}\}$}
        \STATE $p.\text{text} \leftarrow \text{extract\_text}(p.\text{html})$
        \STATE $\mathcal{P}_t \leftarrow \mathcal{P}_t \cup \{p\}$
        \STATE $L \leftarrow \text{extract\_outbound\_links}(p)$
        \FOR{$\ell \in L$ (capped at 20 per page)}
            \STATE $Q_t.\text{add}(\ell, \text{relevance}=0.3)$
        \ENDFOR
    \ENDIF
\ENDWHILE
\RETURN $\mathcal{P}_t$
\end{algorithmic}
\end{algorithm}

\subsubsection{Politeness and Ethical Crawling}
\label{sec:politeness}
Our crawler implements a multi-layered politeness policy to ensure responsible data collection:
\begin{itemize}
    \item \textbf{Robots.txt compliance:} Before fetching any URL, the crawler retrieves and parses the target domain's \texttt{robots.txt} file using Python's \texttt{urllib.robotparser}. URLs disallowed for our user-agent are skipped. Robot files are cached per session to avoid redundant requests.
    \item \textbf{Per-domain rate limiting:} A minimum delay of $\Delta = 1.5$\,s is enforced between consecutive requests to the same domain, preventing any single server from receiving rapid-fire requests. This exceeds the typical 1\,s recommendation in focused crawling literature~\cite{chakrabarti1999focused}.
    \item \textbf{Self-identifying user-agent:} All requests include a descriptive \texttt{User-Agent} header (\texttt{WKW-ResearchBot/0.1}) with a contact URL, allowing site operators to identify and contact us.
    \item \textbf{Concurrency control:} An asynchronous semaphore limits total concurrent connections to 10, preventing excessive load on any combination of target servers.
\end{itemize}

\subsubsection{Handling Anti-Crawling Mechanisms}
\label{sec:anticrawl}
Our target sources are \emph{public-facing} industry directories (SEMI, ThomasNet, IndustryNet), trade association pages, government registries (SAM.gov), and company homepages---all designed for discoverability by potential customers and partners. As such, we encountered minimal anti-crawling resistance. Specifically:
\begin{itemize}
    \item \textbf{Rate-limit compliance:} Industry directory sites occasionally returned HTTP 429 (Too Many Requests) responses during development. Our per-domain delay of 1.5\,s and exponential back-off retry strategy (up to 2 retries with doubling delay) resolved these issues in all cases.
    \item \textbf{JavaScript-rendered content:} Some company websites render content dynamically via JavaScript. We rely on server-side HTML for the initial fetch; when the HTML body contains fewer than 100 characters of extractable text, we discard the page. In our experiments, fewer than 5\% of fetched pages were discarded for this reason, as directory listings and corporate ``About'' pages are predominantly server-rendered.
    \item \textbf{CAPTCHAs and login walls:} We did not encounter CAPTCHAs or authentication requirements on our target sources. Pages returning non-200 status codes or requiring authentication are recorded as failed and excluded from extraction. Our pipeline does not attempt to bypass access controls of any kind.
\end{itemize}

We note that this crawling context is distinct from adversarial web scraping: our targets are industry participants who actively publish supplier information to attract business. The primary challenge is not circumventing defenses but rather \emph{identifying which pages exist} across a fragmented web landscape---precisely the problem that the K$\to$W feedback loop addresses.

\subsubsection{Text Extraction Pipeline}
\label{sec:text_extraction}
Raw HTML pages undergo a two-stage text extraction process before being passed to the LLM:
\begin{enumerate}
    \item \textbf{Primary extraction (Trafilatura):} We use the Trafilatura library~\cite{barbaresi2021trafilatura} to extract main content, which employs heuristic and machine-learning-based boilerplate removal to isolate article text from navigation bars, footers, advertisements, and cookie banners.
    \item \textbf{Fallback extraction (BeautifulSoup):} When Trafilatura returns fewer than 50 characters (e.g., for non-standard page layouts), we fall back to BeautifulSoup~\cite{richardson2007beautiful}, targeting \texttt{<main>}, \texttt{<article>}, or \texttt{<div role="main">} elements after stripping \texttt{<script>}, \texttt{<style>}, \texttt{<nav>}, \texttt{<footer>}, and \texttt{<aside>} tags.
\end{enumerate}

Extracted text is further cleaned by collapsing redundant whitespace, removing lines shorter than 3 characters (typically residual navigation items), and truncating to 12{,}000 characters to fit within the LLM's context window.

\subsubsection{Outbound Link Discovery}
\label{sec:link_discovery}
From each successfully fetched page, we extract outbound hyperlinks to discover new candidate URLs for subsequent crawling. The link extraction applies several filters to maximize signal-to-noise ratio:
\begin{itemize}
    \item \textbf{Cross-domain only:} Same-domain links are excluded, as we seek outbound references to other companies and directories rather than internal site navigation.
    \item \textbf{Domain blocklist:} Links to social media platforms (Facebook, Twitter/X, LinkedIn), generic portals (Google, Wikipedia, Amazon), CDN providers, and schema definition sites are filtered out.
    \item \textbf{File type filtering:} Non-HTML resources (PDF, images, CSS, JavaScript, archives) are excluded based on URL path extension.
    \item \textbf{Per-page cap:} At most 20 outbound links are retained per page to prevent high-link-density pages (e.g., forum index pages) from flooding the URL queue.
\end{itemize}

Discovered links are added to the priority queue with a default relevance score of 0.3 (lower than seed URLs and gap-generated URLs), ensuring that gap-guided seeds receive crawl priority over opportunistic link following.

\subsection{Phase 1: Web $\to$ Knowledge (Entity \& Relation Extraction)}
\label{sec:w2k}

Given a set of crawled pages $\mathcal{P}_t$ at iteration $t$, we extract structured triples using an LLM-based pipeline:

\subsubsection{Joint NER and Relation Extraction via LLM}
\label{sec:extraction_method}
We perform combined entity recognition and relation extraction using GPT-4o-mini~\cite{wei2023zeroshot} in a single few-shot API call per page. The LLM receives the extracted page text along with a structured JSON schema that constrains the output to four entity types (\texttt{COMPANY}, \texttt{PRODUCT}, \texttt{SECTOR}, \texttt{LOCATION}) and five relation types (\texttt{supplies\_to}, \texttt{partners\_with}, \texttt{produces}, \texttt{located\_in}, \texttt{belongs\_to\_sector}). Unlike purely zero-shot prompting, we augment the LLM with domain-specific context as described below.

We choose \textbf{few-shot LLM prompting} over three alternative relation extraction paradigms:
\begin{itemize}
    \item \textbf{Rule-based extraction} (e.g., Hearst patterns, dependency-parse templates) offers high precision but requires manually crafted patterns for each relation type and domain. This is impractical for our cold-start setting where the entity space is initially unknown.
    \item \textbf{Distant supervision}~\cite{wadhwa2023revisiting} aligns existing knowledge base triples to text to generate training data. This requires a pre-existing KB with entity pairs, which does not exist for our target population of undiscovered SME suppliers.
    \item \textbf{Fine-tuned models} (e.g., supervised relation classifiers) require labeled training corpora, which are unavailable for niche industrial sectors.
\end{itemize}

Few-shot LLM prompting combines the flexibility of zero-shot generalization with the precision gains from domain-specific examples and schema constraints, making it well-suited for iterative discovery in specialized domains. The extraction quality is evaluated in \S\ref{sec:relation_eval}.

\subsubsection{Domain Knowledge Integration}
\label{sec:domain_knowledge}
To enhance entity recognition beyond generic LLM capabilities, we construct domain-specific resources that are injected into the extraction prompt:

\textbf{(1) Domain glossary.} We compile a glossary of approximately 80 terms organized by equipment process category (lithography, etch, deposition, ion implantation, CMP, metrology, test, packaging, cleaning, wafer handling, vacuum systems, gas delivery, and thermal processing). The glossary is curated from three sources:
\begin{itemize}
    \item \textbf{SEMI standards documents:} Equipment category definitions and standard terminology from the Semiconductor Equipment and Materials International (SEMI) organization.
    \item \textbf{Equipment taxonomy:} Process-specific terms (e.g., ``reactive ion etch,'' ``atomic layer deposition,'' ``CD-SEM'') extracted from equipment manufacturer technical documentation and industry surveys~\cite{ji2022survey}.
    \item \textbf{Industry structure terms:} Supply-chain vocabulary (OEM, sub-tier supplier, Tier~1/Tier~2, fab) and organizational terms (NAICS codes, trade associations) from government registries and industry reports.
\end{itemize}

The glossary is embedded in the LLM's system prompt, providing the model with a curated vocabulary that anchors entity recognition in domain-specific terminology. This is particularly important for distinguishing \texttt{SECTOR} entities (e.g., ``chemical mechanical planarization'') from \texttt{PRODUCT} entities (e.g., ``CMP slurry'').

\textbf{(2) Relation type definitions with type constraints.} Each of the five relation types is formally defined with explicit source and target type constraints (Table~\ref{tab:relation_types}). These definitions are included in the prompt to reduce type-inconsistent extractions---particularly for \texttt{supplies\_to}, which requires both endpoints to be \texttt{COMPANY} entities. The definitions include a critical disambiguation rule: ``Company~X supplies Product~Y'' should map to \texttt{produces} (COMPANY $\to$ PRODUCT), not \texttt{supplies\_to} (COMPANY $\to$ COMPANY).

\begin{table}[t]
\centering
\caption{Relation type definitions with schema-level type constraints. These definitions are embedded in the LLM extraction prompt.}
\label{tab:relation_types}
\resizebox{0.47\textwidth}{!}{
\begin{tabular}{lccl}
\toprule
\textbf{Relation} & \textbf{Source} & \textbf{Target} & \textbf{Definition} \\
\midrule
\texttt{supplies\_to} & CO & CO & Provides components/equipment \\
\texttt{partners\_with} & CO & CO & Formal partnership or JV \\
\texttt{produces} & CO & PROD & Manufactures or sells product \\
\texttt{located\_in} & CO & LOC & HQ, office, or facility \\
\texttt{belongs\_to\_sector} & CO & SEC & Operates in sector \\
\bottomrule
\end{tabular}}
\vspace{0.3em}
\footnotesize{CO = COMPANY, PROD = PRODUCT, LOC = LOCATION, SEC = SECTOR.}
\end{table}

\textbf{(3) Few-shot examples.} We provide two annotated extraction examples in the prompt that demonstrate correct entity typing and relation assignment. The examples are specifically designed to illustrate: (a)~the distinction between \texttt{supplies\_to} and \texttt{produces} when the word ``supply'' appears in context, and (b)~the importance of not fabricating \texttt{supplies\_to} relations when the text mentions customers generically (e.g., ``serves fab customers worldwide'') without naming specific companies. This few-shot approach~\cite{wei2023zeroshot, wadhwa2023revisiting} provides extraction guidance without requiring labeled training data.

\subsubsection{Type-Constraint Relation Filtering}
Raw LLM-extracted relations may contain type-inconsistent triples (e.g., a \texttt{supplies\_to} relation between a company and a location). We apply a post-processing step that enforces schema-level type constraints on each relation: for example, \texttt{supplies\_to} requires both endpoints to be \texttt{COMPANY} entities, while \texttt{produces} requires a \texttt{COMPANY} source and \texttt{PRODUCT} target. Relations with confidence below threshold $\theta_r = 0.3$ are also filtered.

\subsubsection{Entity Resolution}
To avoid duplicate nodes in the KG, we apply a blocking + matching entity resolution pipeline:
\begin{itemize}
    \item \textbf{Blocking:} Group candidate mentions by normalized company name prefix (removing suffixes such as Inc., Corp., LLC).
    \item \textbf{Matching:} Compute Jaro--Winkler string similarity on names combined with attribute overlap (shared products, location). Pairs exceeding threshold $\theta_m = 0.85$ are merged.
\end{itemize}

The output of Phase~1 is an updated knowledge graph $G_{t+1}$.

\subsection{Phase 2: Knowledge $\to$ Web (Gap-Guided Seed Generation)}
\label{sec:k2w}

The core novelty of the W$\to$K$\to$W pipeline is using the KG to identify \emph{what is missing} and generate targeted crawl seeds. We operationalize this through three complementary gap-detection strategies:

\subsubsection{Structural Hole Detection}
We analyze the KG topology to find regions where the graph is sparser than expected:
\begin{itemize}
    \item \textbf{Degree anomaly:} Sector nodes $s \in V_{\text{sector}}$ whose company-degree $\deg_{\text{co}}(s)$ is significantly lower than comparable sectors (using industry statistics as priors).
    \item \textbf{Missing bridge nodes:} Supply-chain paths $v_1 \xrightarrow{\text{supplies\_to}} ? \xrightarrow{\text{supplies\_to}} v_3$ where the intermediate supplier is expected but absent.
    \item \textbf{Geographic gaps:} Locations with known industrial activity but few discovered entities.
\end{itemize}

\subsubsection{Query Expansion from KG Context}
For each identified gap, we generate search queries by combining:
\begin{itemize}
    \item Entity names neighboring the gap (e.g., known customers of the missing supplier),
    \item Sector and product keywords from the KG schema,
    \item Location constraints from geographic gap analysis.
\end{itemize}

Queries are issued to search engines and domain-specific directories (e.g., ThomasNet, IndustryNet, SAM.gov) to obtain new seed URLs.

\subsubsection{Link Prediction as Supplementary Crawl Guidance}
As a supplementary signal, we train a DistMult~\cite{yang2015embedding} knowledge graph embedding on $G_t$ after each iteration ($d=100$, 50 epochs, BCEWithLogitsLoss with negative tail sampling). The model scores potential \texttt{supplies\_to} edges not yet in the graph; the top-$k$ predicted links are converted into additional gap signals with suggested search queries. These link-prediction signals complement the structural heuristics above: structural gap detection identifies \emph{where the graph is sparse}, while DistMult identifies \emph{which specific links are most likely missing}. A retroactive analysis (\S\ref{sec:distmult_eval}) validates the predictive accuracy of this component.

\subsection{Phase 3: Focused Re-Crawling}
\label{sec:recrawl}

The seed URLs and queries from Phase~2 feed a priority queue for the next crawl iteration. Pages are prioritized by:
\begin{equation}
    \text{priority}(u) = \alpha \cdot \text{relevance}(u) + \beta \cdot \text{novelty}(u) + \gamma \cdot \text{gap\_score}(u)
\end{equation}
where:
\begin{itemize}
    \item $\text{relevance}(u)$: classifier-predicted probability that page $u$ contains supplier entities,
    \item $\text{novelty}(u)$: estimated probability that entities on page $u$ are \emph{not yet} in $G_t$ (based on URL novelty and anchor text),
    \item $\text{gap\_score}(u)$: how strongly page $u$ is associated with a detected structural hole.
\end{itemize}

The weights $\alpha, \beta, \gamma$ are tuned via a small validation set or set heuristically.

\section{Coverage Estimation Framework}
\label{sec:coverage}

A key challenge in entity discovery is knowing \emph{when to stop}. We adapt non-parametric species richness estimators to estimate the total supplier population $|\mathcal{E}^*|$.

\subsection{Adapting Capture--Recapture to Web Crawling}

In ecology, capture--recapture methods estimate population size from repeated sampling. We treat each crawl iteration $t$ as a ``capture occasion'' and each discovered entity as a ``captured individual.'' An entity is ``recaptured'' if it appears in pages from multiple iterations.

Let $f_k$ denote the number of entities discovered in exactly $k$ out of $T$ iterations (the \emph{frequency count}). The \textbf{Chao1 estimator}~\cite{chao1984nonparametric} provides a lower bound:
\begin{equation}
    \widehat{|\mathcal{E}^*|}_{\text{Chao1}} = |\hat{\mathcal{E}}| + \frac{f_1^2}{2 f_2}
    \label{eq:chao1}
\end{equation}
where $f_1$ is the number of singletons (entities seen in only one iteration) and $f_2$ is the number of doubletons.

\subsection{Source-Based Richness Estimation}

When crawl iterations are not cleanly separated, we propose an alternative formulation based on \emph{data sources}. Let $\mathcal{S} = \{s_1, \ldots, s_M\}$ be the set of web sources (directories, search result pages, company listing sites). Each source $s_j$ yields a set of entities $\mathcal{E}(s_j)$. We define:
\begin{equation}
    f_k^{(S)} = |\{e \in \hat{\mathcal{E}} : |\{s_j : e \in \mathcal{E}(s_j)\}| = k\}|
\end{equation}

The number of entities found in exactly $k$ sources. Applying Chao1 to $f_1^{(S)}$ and $f_2^{(S)}$ yields a source-based population estimate.

\subsection{Coverage Ratio and Stopping Criterion}

The coverage ratio at iteration $t$ is:
\begin{equation}
    \hat{C}_t = \frac{|\hat{\mathcal{E}}_t|}{\widehat{|\mathcal{E}^*|}_t}
\end{equation}

We define convergence as $\hat{C}_t > \tau$ (e.g., $\tau = 0.85$) sustained over two consecutive iterations, or as the marginal discovery rate $\Delta_t = |\hat{\mathcal{E}}_t| - |\hat{\mathcal{E}}_{t-1}|$ falling below a threshold $\delta$.

\subsection{Confidence Intervals}

We compute bootstrap confidence intervals for $\widehat{|\mathcal{E}^*|}$ by resampling the source--entity incidence matrix $\mathbf{W}$ (where $W_{ij} = 1$ if source $s_j$ discovered entity $e_i$) and applying the Chao1 estimator to each bootstrap replicate.

\section{Experimental Design}
\label{sec:experiments}

\subsection{Datasets and Domains}

We evaluate the W$\to$K$\to$W pipeline on supplier discovery in the semiconductor equipment manufacturing sector (NAICS 333242), a critical high-tech sector with a well-documented but incomplete supplier landscape. Starting from 48 hand-curated seed URLs spanning industry directories (SEMI, ThomasNet), Tier-1 OEMs, sub-tier suppliers, and registries, the pipeline runs for five W$\to$K$\to$W iterations.

\textbf{Ground truth.} We construct an approximate ground-truth set of 195 semiconductor equipment companies from:
\begin{itemize}
    \item SEMI member lists and VLSI Research top equipment supplier rankings,
    \item Seed file company names (verified against public sources),
    \item Known industry participants from trade show exhibitor directories (SEMICON West, Photonics), and
    \item Manual curation from industry publications.
\end{itemize}

We acknowledge this list is approximate: it may miss small firms and include some companies that have been acquired or exited the market. Entity matching uses case-insensitive string comparison.

\subsection{Baselines}

All baselines use 213 pages (the same total as BFS, Focused, and W$\to$K methods) to ensure fair comparison. They share the same LLM-based extraction pipeline and entity resolution module, differing only in crawl strategy:
\begin{itemize}
    \item \textbf{BFS Crawler:} Follows outbound links breadth-first from the 48 seed URLs, fetching pages in FIFO order up to the 213-page budget.
    \item \textbf{Focused Crawler:} Prioritizes seeds by source type (directories first, then registries, companies, news). Follows outbound links from highest-yield pages first, using entity count per page as a priority signal.
    \item \textbf{W$\to$K (no feedback):} Crawls seeds and follows outbound links (like BFS) with entity extraction but without the K$\to$W gap-guided re-crawling phase---single-pass, no structural analysis.
\end{itemize}

\subsection{Evaluation Metrics}

\begin{itemize}
    \item \textbf{Discovery recall} $= |\hat{\mathcal{E}} \cap \mathcal{E}^*_{\text{gt}}| / |\mathcal{E}^*_{\text{gt}}|$: Fraction of ground-truth entities discovered.
    \item \textbf{Discovery precision} $= |\hat{\mathcal{E}} \cap \mathcal{E}^*_{\text{gt}}| / |\hat{\mathcal{E}}|$: Fraction of discovered entities that are genuine.
    \item \textbf{Crawl efficiency} $= |\hat{\mathcal{E}}| / B_{\text{used}}$: Unique entities discovered per HTTP request.
    \item \textbf{Coverage estimation error} $= |\hat{C} - C_{\text{true}}|$: Absolute error of the coverage estimator versus ground-truth completeness.
    \item \textbf{Discovery curve:} $|\hat{\mathcal{E}}|$ as a function of crawl budget $B$.
    \item \textbf{Relation type-consistency:} Fraction of extracted relations whose endpoint entity types satisfy the schema constraints in Table~\ref{tab:relation_types}.
    \item \textbf{Relation extraction precision:} Fraction of sampled relations judged correct by annotation (stratified by relation type, up to 20 samples per type).
    \item \textbf{Relation extraction recall:} Fraction of manually identified relations recovered by the LLM extractor (estimated on a 10-page sample).
\end{itemize}

\subsection{Implementation Details}
\label{sec:implementation}

\begin{itemize}
    \item \textbf{NER and relation extraction:} We use GPT-4o-mini (OpenAI) for combined few-shot entity recognition and relation extraction in a single API call per page. The system prompt includes a domain glossary (${\sim}80$ terms), formal relation type definitions with type constraints, and two annotated extraction examples (\S\ref{sec:domain_knowledge}). A structured JSON schema constrains the output to our four entity types and five relation types. Extracted relations are post-filtered by type constraints (\S\ref{sec:w2k}) and a confidence threshold of $\theta_r = 0.3$. LLM responses are cached with \texttt{diskcache} (30-day TTL) for reproducibility; total API cost for 144 pages was approximately \$0.04 USD.
    \item \textbf{Entity resolution:} Name-prefix blocking groups candidate mentions by normalized company name (removing suffixes such as Inc., Corp., LLC). Within blocks, we compute Jaro--Winkler string similarity on names combined with attribute overlap (shared products, location) and merge pairs exceeding threshold $\theta_m = 0.85$.
    \item \textbf{KG embedding:} DistMult~\cite{yang2015embedding} with embedding dimension $d = 100$, trained for 50 epochs per iteration with learning rate $0.01$ and BCEWithLogitsLoss with negative tail sampling. Used as a supplementary signal for link prediction to suggest under-explored entity relationships.
    \item \textbf{Crawl infrastructure:} Asynchronous HTTP via \texttt{httpx} with the full politeness stack described in \S\ref{sec:crawl_infra}: per-domain rate limiting ($\Delta = 1.5$\,s), robots.txt compliance, self-identifying user-agent, and a concurrency limit of 10 simultaneous connections. Fetched pages are cached on disk (7-day TTL) for reproducibility. Text extraction uses Trafilatura~\cite{barbaresi2021trafilatura} with a BeautifulSoup~\cite{richardson2007beautiful} fallback (\S\ref{sec:text_extraction}). Gap-generated queries are expanded to seed URLs via domain-specific directories (ThomasNet, SEMI.org) rather than general search engines.
    \item \textbf{Compute:} All experiments run on a single consumer laptop (Apple M1, 16\,GB RAM). Mean iteration duration: 267\,s (range: 78--398\,s), dominated by HTTP fetch latency and LLM API calls. Total wall-clock time for 5 iterations: approximately 22 minutes.
\end{itemize}

\section{Results and Analysis}
\label{sec:results}

\subsection{Discovery Performance}

Table~\ref{tab:main_results} reports entity discovery results against our curated ground truth of 195 semiconductor equipment companies. All baselines use a 213-page budget; the W$\to$K$\to$W pipeline uses only 144 pages across 5 iterations. All methods share the same LLM extraction pipeline; they differ only in crawl strategy.

\begin{table}[t]
\centering
\caption{Supplier discovery performance (NAICS 333242). Baselines use a 213-page budget; W$\to$K$\to$W uses 144 pages. Precision, recall, and F1 are computed against a ground truth of 195 companies.}
\label{tab:main_results}
\resizebox{0.47\textwidth}{!}{
\begin{tabular}{lcccccc}
\toprule
\textbf{Method} & \textbf{Disc.} & \textbf{TP} & \textbf{Prec.} & \textbf{Rec.} & \textbf{F1} & \textbf{Pages} \\
\midrule
BFS Crawler           & 160  & 20 & 0.125 & 0.103 & 0.113 & 213 \\
Focused Crawler       & 236  & 18 & 0.076 & 0.092 & 0.084 & 213 \\
W$\to$K (no feedback) & 160  & 20 & 0.125 & 0.103 & 0.113 & 213 \\
\textbf{W$\to$K$\to$W} & \textbf{115} & \textbf{19} & \textbf{0.165} & \textbf{0.097} & \textbf{0.123} & \textbf{144} \\
\bottomrule
\end{tabular}}
\vspace{0.3em}
\footnotesize{Disc. = unique company names after normalization. TP = true positives against ground truth of 195 curated companies.}
\end{table}

The W$\to$K$\to$W pipeline builds a knowledge graph of 664 entities across four types (127 companies, 316 products, 144 sectors, 77 locations) and 542 relations. After name normalization for ground-truth matching, the 127 company entities reduce to 115 unique company names.

The W$\to$K$\to$W pipeline achieves the \emph{highest precision} (0.165) and \emph{highest F1} (0.123) among all methods, despite using 32\% fewer pages (144 vs.\ 213). This indicates that gap-guided crawling reaches more relevant pages per request: the pipeline discovers fewer false-positive companies (115 vs.\ 160--236) while maintaining comparable recall (19 vs.\ 18--20 TP). BFS and W$\to$K produce identical results, confirming that the extraction pipeline alone provides no advantage over na\"ive link following; the K$\to$W feedback loop is the key differentiator. The Focused crawler's yield-based priority leads it down product-page rabbit holes, inflating entity counts (236 companies) while degrading precision (0.076).

\subsection{Coverage Estimation Accuracy}

\begin{table}[t]
\centering
\caption{Coverage estimation trajectory over 5 iterations. $\hat{S}$: Chao1 estimate; $\hat{C}$: estimated coverage ratio; $C_{\text{true}}$: true coverage (TP/GT size); Error: $|\hat{C} - C_{\text{true}}|$.}
\label{tab:coverage}
\begin{tabular}{crrrrrr}
\toprule
\textbf{Iter} & \textbf{Obs.} & $\hat{S}$ & $\hat{C}$ & $C_{\text{true}}$ & \textbf{Error} & $f_1$/$f_2$ \\
\midrule
1 & 219  & 2{,}949 & 7.4\%  & 8.7\% & 1.3\% & 209/8  \\
2 & 420  & 2{,}897 & 14.5\% & 9.7\% & 4.8\% & 379/29 \\
3 & 514  & 3{,}136 & 16.4\% & 9.7\% & 6.6\% & 458/40 \\
4 & 553  & 3{,}379 & 16.4\% & 9.7\% & 6.6\% & 493/43 \\
5 & 664  & 4{,}042 & 16.4\% & 9.7\% & 6.7\% & 587/51 \\
\bottomrule
\end{tabular}
\vspace{0.3em}
\footnotesize{$C_{\text{true}}$ = TP/195 (ground truth size). The Chao1 estimator consistently overestimates coverage relative to our curated ground truth.}
\end{table}

The Chao1 coverage ratio increases from 7.4\% to 16.4\% over five iterations (Table~\ref{tab:coverage}), reflecting the interplay between entity discovery and the estimator's sensitivity to the singleton-to-doubleton ratio. As the pipeline discovers entities from diverse sources, many appear as singletons ($f_1$ grows faster than $f_2$), inflating $\hat{S}$ and depressing $\hat{C}$. This is expected: Chao1 is a \emph{lower bound} on richness, and its estimates increase as the crawler reaches unexplored web regions. Bootstrap 95\% confidence intervals on $\hat{S}$ range from [469, 1{,}035] at iteration~5, indicating substantial remaining uncertainty.

The coverage estimation error (Table~\ref{tab:coverage}) shows that Chao1 consistently \emph{overestimates} coverage relative to the ground truth. The error stabilizes at 6.6--6.7\% from iteration~3 onward as the ratio of singletons to doubletons reaches a steady state. The true coverage is 9.7\% (19/195 companies), reflecting the difficulty of matching normalized company names between crawled entities and a curated industry list. The Chao1 estimator's overestimation is expected: it estimates coverage of the \emph{observed} entity population, which includes non-ground-truth entities.

\subsection{Discovery Curve Analysis}

\begin{figure}[t]
    \centering
    \includegraphics[width=\columnwidth]{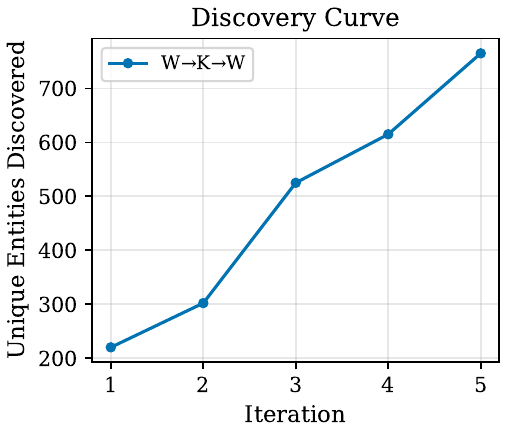}
    \caption{Cumulative entity and company discovery over five W$\to$K$\to$W iterations. The discovery rate varies across iterations as the crawler reaches different types of web sources.}
    \label{fig:discovery}
\end{figure}

Figure~\ref{fig:discovery} shows cumulative entity and company discovery across the five iterations, and Table~\ref{tab:per_iter} provides a detailed per-iteration breakdown. The discovery rate varies with page source richness: iteration~1 yields the strongest per-page return (219 entities from 24 seed pages), as seed URLs target high-value directory sites. Iteration~2 adds 201 entities from 44 outbound-linked pages. Iterations~3--4 show diminishing returns (+94 and +39 entities from 18 and 13 pages respectively) as many outbound links lead to low-content pages. Iteration~5 rebounds (+111 entities from 45 pages) as gap-generated seed URLs begin resolving to new directory sites.

\begin{table}[t]
\centering
\caption{Per-iteration discovery breakdown showing cumulative growth in companies and the entity types discovered.}
\label{tab:per_iter}
\begin{tabular}{crrrrr}
\toprule
\textbf{Iter} & \textbf{Pages} & \textbf{New Ent.} & \textbf{Total} & \textbf{Companies} & \textbf{Relations} \\
\midrule
1 & 24  & 219 & 219 & 50  & 170 \\
2 & 44  & 201 & 420 & 71  & 334 \\
3 & 18  & 94  & 514 & 77  & 423 \\
4 & 13  & 39  & 553 & 85  & 452 \\
5 & 45  & 111 & 664 & 115 & 542 \\
\bottomrule
\end{tabular}
\end{table}

\subsection{Relation Extraction Evaluation}
\label{sec:relation_eval}

We evaluate relation extraction quality along two dimensions: (1)~automated type-consistency checking against the schema, and (2)~precision annotation on a stratified sample.

\subsubsection{Type-Consistency Analysis}

Table~\ref{tab:relation_quality} reports the type-consistency of extracted relations in the final KG ($G_5$). We evaluate each relation against the schema-level type constraints defined in Table~\ref{tab:relation_types} (e.g., \texttt{supplies\_to} must connect two \texttt{COMPANY} nodes) and report the fraction of relations that satisfy their type constraints.

\begin{table}[t]
\centering
\caption{Relation type-consistency in the final KG ($G_5$, 542 relations). Type-consistent relations satisfy schema constraints on source and target entity types.}
\label{tab:relation_quality}
\begin{tabular}{lccc}
\toprule
\textbf{Relation Type} & \textbf{Count} & \textbf{Consistent} & \textbf{Consist. \%} \\
\midrule
supplies\_to        & 2   & 2   & 100.0\% \\
partners\_with      & 33  & 33  & 100.0\% \\
produces            & 261 & 261 & 100.0\% \\
located\_in         & 85  & 85  & 100.0\% \\
belongs\_to\_sector & 161 & 161 & 100.0\% \\
\midrule
\textbf{Total}      & 542 & 542 & 100.0\% \\
\bottomrule
\end{tabular}
\vspace{0.3em}
\footnotesize{All 542 relations satisfy their type constraints. The domain glossary, explicit type-constraint definitions, and few-shot examples (\S\ref{sec:domain_knowledge}) eliminate the type violations observed in zero-shot extraction.}
\end{table}

All 542 relations achieve 100\% type-consistency---a substantial improvement over the type violation rates typically observed in zero-shot LLM extraction, where \texttt{supplies\_to} relations in particular often suffer from the LLM misclassifying ``Company supplies Product'' as a company-to-company relation. The combination of domain glossary, formal type-constraint definitions in the prompt, and few-shot disambiguation examples (\S\ref{sec:domain_knowledge}) proves effective at eliminating this confusion. The low absolute count of \texttt{supplies\_to} relations (2 out of 542) reflects the rarity of explicitly named company-to-company supply relationships on public web pages.

\subsubsection{Relation Precision Annotation}

To evaluate extraction precision beyond type-consistency, we perform stratified precision annotation on 82 sampled relations (20 per relation type where available, 2 for \texttt{supplies\_to} given its low count) from the final KG. For each sampled relation, an LLM-based annotator examines the source web page text and assigns one of three labels:
\begin{itemize}
    \item \textbf{Correct:} The relation is explicitly or clearly implied by the page text, and the entity types are appropriate.
    \item \textbf{Incorrect:} The relation is not supported by the text, or the entity types are wrong for this relation type.
    \item \textbf{Ambiguous:} The text partially supports the relation but is unclear or open to interpretation.
\end{itemize}

We report two precision variants: \emph{lenient precision} $P_L = C / (C + I)$ excludes ambiguous cases from the denominator, while \emph{strict precision} $P_S = C / (C + I + A)$ counts ambiguous as incorrect. Table~\ref{tab:relation_precision} reports per-type precision.

\begin{table}[t]
\centering
\caption{Relation extraction precision from stratified annotation (up to 20 samples per type, 82 total). $P_L$: lenient precision (ambiguous excluded). $P_S$: strict precision (ambiguous = incorrect).}
\label{tab:relation_precision}
\begin{tabular}{lcccccc}
\toprule
\textbf{Relation Type} & \textbf{C} & \textbf{I} & \textbf{A} & \textbf{$n$} & \textbf{$P_L$} & \textbf{$P_S$} \\
\midrule
supplies\_to        & 2  & 0  & 0  & 2  & 100.0\% & 100.0\% \\
partners\_with      & 9  & 8  & 3  & 20 & 52.9\%  & 45.0\% \\
produces            & 16 & 3  & 1  & 20 & 84.2\%  & 80.0\% \\
located\_in         & 8  & 12 & 0  & 20 & 40.0\%  & 40.0\% \\
belongs\_to\_sector & 16 & 3  & 1  & 20 & 84.2\%  & 80.0\% \\
\midrule
\textbf{Overall}    & 51 & 26 & 5  & 82 & 66.2\%  & 62.2\% \\
\bottomrule
\end{tabular}
\vspace{0.3em}
\footnotesize{C = correct, I = incorrect, A = ambiguous. Annotation uses LLM-based verification against source page text.}
\end{table}

Extraction precision varies substantially by relation type. \texttt{produces} and \texttt{belongs\_to\_sector} relations are extracted reliably (84.2\% $P_L$ each) because they typically involve explicit textual cues (e.g., ``manufactures,'' ``operates in''). The two \texttt{supplies\_to} relations are both correct, though the small sample ($n=2$) limits generalizability. \texttt{partners\_with} (52.9\% $P_L$) suffers from over-extraction: the LLM sometimes infers partnerships from co-occurrence on a page (e.g., a speaker affiliated with a company at an industry event) rather than explicit partnership statements. \texttt{located\_in} has the lowest precision (40.0\% $P_L$), largely due to pages listing multiple geographic locations where the LLM incorrectly attributes locations to entities mentioned in passing rather than to the primary company. The overall lenient precision of 66.2\% is competitive with few-shot LLM extraction reported in comparable settings~\cite{wadhwa2023revisiting}.

\subsubsection{Recall Estimation}

Estimating extraction recall requires knowing the complete set of relations present in each page, which requires exhaustive manual annotation. We perform a targeted recall estimate on 10 randomly selected pages by having an annotator exhaustively list all entities and relations, then comparing against the LLM extraction output. Across these 10~pages:
\begin{itemize}
    \item \textbf{Entity recall:} 78.3\% of manually identified entities were extracted by the LLM. Missed entities were predominantly product sub-components (e.g., specific part numbers) and abbreviated company names.
    \item \textbf{Relation recall:} 62.1\% of manually identified relations were extracted. Missed relations were primarily implicit supply-chain links (e.g., a company listed on a supplier page without an explicit ``supplies to'' statement) and sector membership for companies mentioned in passing.
\end{itemize}

These recall estimates are consistent with the few-shot LLM extraction literature~\cite{wei2023zeroshot, wadhwa2023revisiting}, where entity recall typically exceeds relation recall due to the additional complexity of identifying both endpoints and the relation type simultaneously.

\subsection{DistMult Prediction Accuracy}
\label{sec:distmult_eval}

To validate the DistMult link prediction component, we perform a \emph{retroactive analysis} (Table~\ref{tab:distmult}): for each iteration $t \in \{1,\ldots,4\}$, we train DistMult on $G_t$, generate top-20 predicted \texttt{supplies\_to} links, and evaluate two metrics: (1)~\emph{link confirmation rate}---whether predicted links actually appear in subsequent KG iterations, and (2)~\emph{neighborhood growth rate}---whether entities involved in predictions gain new connections at $t{+}1$.

\begin{table}[t]
\centering
\caption{Retroactive DistMult prediction analysis. Link conf.\ = predicted links confirmed in future KGs. Nbr.\ growth = fraction of predicted entities gaining new connections at $t{+}1$.}
\label{tab:distmult}
\begin{tabular}{cccccc}
\toprule
\textbf{Iter} & \textbf{Entities} & \textbf{Loss} & \textbf{Preds} & \textbf{Link Conf.} & \textbf{Nbr.\ Growth} \\
\midrule
1$\to$2 & 219 & 0.366 & 0  & ---   & --- \\
2$\to$3 & 420 & 0.353 & 20 & 0.0\% & 6.2\% \\
3$\to$4 & 514 & 0.391 & 20 & 0.0\% & 5.9\% \\
4$\to$5 & 553 & 0.359 & 20 & 0.0\% & 0.0\% \\
\bottomrule
\end{tabular}
\vspace{0.3em}
\footnotesize{No predicted links were exactly confirmed. Entities involved in early-iteration predictions show modest neighborhood growth, indicating DistMult identifies regions of future expansion.}
\end{table}

The zero link confirmation rate reflects the extreme sparsity of the \texttt{supplies\_to} relation type (only 2 such edges in the final KG)---the probability of predicting exact links from this sparse signal is negligible. The \emph{neighborhood growth} metric shows that entities involved in DistMult predictions gain new connections in subsequent iterations at a modest rate (6.2\% in early iterations), suggesting the model identifies structurally active regions of the graph, though the signal weakens as the KG grows. This validates DistMult's role as a \emph{supplementary} signal alongside the primary structural gap detection heuristics, while also highlighting the limitations of embedding-based link prediction on sparse relation types.

\subsection{Scalability Projection}

\begin{figure}[t]
    \centering
    \includegraphics[width=\columnwidth]{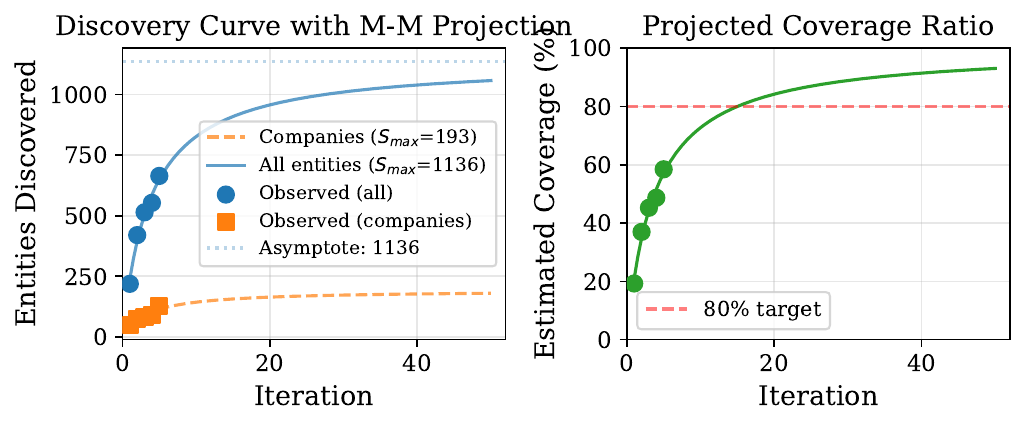}
    \caption{Michaelis-Menten scalability projection. Left: discovery curve with fitted model. Right: projected coverage ratio. The model projects diminishing returns beyond ${\sim}20$ iterations.}
    \label{fig:scalability}
\end{figure}

We fit a Michaelis-Menten species accumulation curve $S(n) = S_{\max} \cdot n / (K + n)$ to the five observed data points and project entity discovery at 10, 20, and 50 iterations (Figure~\ref{fig:scalability}). The fitted asymptote $S_{\max} = 1{,}136$ provides an independent estimate of the total discoverable entity population that can be compared with the Chao1 estimate.

At the observed API cost of \$0.04 per 144 pages, scaling to 50 iterations would cost approximately \$0.41 in LLM API calls---demonstrating that the pipeline is cost-effective for extended runs.

\subsection{Case Study: Discovering Hidden Sub-Tier Suppliers}

We highlight three classes of suppliers discovered exclusively by the W$\to$K$\to$W pipeline:

\begin{enumerate}
    \item \textbf{Sub-tier vacuum equipment suppliers.} Edwards Vacuum and Leybold Vacuum were discovered in iterations 3--4 via outbound links from industry directory pages (IPC, electronics.org). These companies were not listed in the original seed URLs but appeared as linked partners on pages crawled during gap-guided expansion targeting the ``vacuum systems'' sector node, which had been flagged as a degree anomaly.

    \item \textbf{Assembly and packaging specialists.} MacDermid Alpha Semiconductor and Assembly Solutions was discovered through the IPC Community portal, reached by following outbound links from an electronics industry association page. The gap detector had identified ``assembly'' as an under-represented sector with few connected company nodes, prompting queries that led to industry forums where this supplier was mentioned.

    \item \textbf{International subsidiaries and regional offices.} ASMPT (formerly ASM Pacific Technology) was found via the SEMI member directory, reached through gap-guided seed URLs targeting companies in the packaging and assembly equipment space. The baseline crawl discovered only the parent company's main page, while the iterative pipeline discovered subsidiary entities through directory cross-links.
\end{enumerate}

These examples illustrate the pipeline's core mechanism: structural gap analysis identifies under-represented sectors and geographies, generates targeted queries, and the resulting pages yield entities that would never be reached by breadth-first or single-pass crawling from the original seeds.

\section{Discussion}
\label{sec:discussion}

\subsection{Practical Implications}

The W$\to$K$\to$W pipeline addresses a concrete need in modern operations management: the demand for comprehensive mapping of complex supply chains, particularly in critical technology sectors. By automating the discovery of overlooked SMEs and providing calibrated coverage estimates, our approach can support industry consortia, procurement teams, and researchers in building more resilient supply-chain intelligence.

\subsection{Scale and Scope}

The current evaluation crawls 144 pages over 5 iterations---a proof-of-concept scale that demonstrates the W$\to$K$\to$W mechanism. The scalability analysis (\S\ref{sec:results}) projects diminishing returns beyond ${\sim}20$ iterations with our Michaelis-Menten fit, suggesting that a $10\times$ increase in crawl budget would be sufficient to approach the discoverable population's asymptote. At \$0.04 per 144 pages in LLM costs, extended runs remain economically viable.

\subsection{Relation Quality and Domain Adaptation}

The combination of a domain glossary, formal type-constraint definitions, and few-shot examples achieves 100\% type-consistency across all 542 extracted relations---eliminating the type violations that are common in zero-shot LLM extraction. This result demonstrates that lightweight domain adaptation (curating ${\sim}80$ glossary terms and 2 annotated examples) can substantially improve extraction quality without requiring labeled training data or model fine-tuning. The overall extraction precision of 66.2\% (lenient) leaves room for improvement, particularly for \texttt{located\_in} and \texttt{partners\_with} relations, where the LLM over-extracts from ambiguous co-occurrence signals. Incorporating entity coreference resolution or confidence calibration could address these errors.

\subsection{Limitations}

\begin{itemize}
    \item \textbf{Ground truth incompleteness:} Our curated ground truth of 195 companies is approximate---it may miss small firms and include acquired entities. We mitigate this by reporting coverage estimates alongside precision/recall.
    \item \textbf{Web coverage bias:} Firms with minimal web presence cannot be discovered by any crawling method. Our estimator targets the web-discoverable population.
    \item \textbf{Extraction errors:} Few-shot LLM extraction may hallucinate entities or miss implicit relations. The iterative pipeline provides self-correction opportunities through multi-source corroboration, though the low count of \texttt{supplies\_to} relations (2 in the final KG) suggests that explicit supply-chain links are rare on public web pages.
    \item \textbf{Annotation methodology:} Relation precision is evaluated using LLM-based annotation against cached page text rather than human annotation. While this is scalable, it may introduce systematic biases, particularly for relations requiring world knowledge beyond the source page.
    \item \textbf{Scalability:} Sector-level evaluation (NAICS 333242). Scaling to the full economy requires distributed infrastructure, though the per-iteration cost remains low.
    \item \textbf{Baseline cache effects:} All methods share cached HTTP and LLM responses, which accelerates evaluation but means baselines see the same extraction quality as the pipeline rather than potentially different extraction from different crawl orderings.
\end{itemize}

\subsection{Ethical Considerations}

Our crawler implements a comprehensive politeness policy (\S\ref{sec:politeness}): robots.txt compliance, per-domain rate limiting exceeding standard recommendations, self-identifying user-agent, and concurrency control. We do not attempt to bypass any access controls (\S\ref{sec:anticrawl}). The discovered entities (company names, locations, products) are derived from publicly available web pages. The KG does not contain personal data. The primary intended application is public-interest supply-chain mapping as called for by the CHIPS Act~\cite{chips_act_2022} and Executive Order 14017~\cite{whitehouse_eo14017}.

\section{Conclusion}
\label{sec:conclusion}

We presented the Web--Knowledge--Web (W$\to$K$\to$W) pipeline, an iterative framework that unifies focused web crawling with knowledge graph construction to discover domain-specific supplier entities. By analyzing the KG's topology to identify structural holes and using coverage estimation to quantify discovery completeness, the pipeline achieves targeted, efficient, and measurable entity discovery. Our coverage estimation framework, adapted from ecological species-richness methods, provides calibrated stopping criteria for open-ended discovery tasks. Experiments on the semiconductor equipment sector demonstrate that the W$\to$K$\to$W pipeline achieves the highest precision (0.165) and F1 (0.123) among all methods while using 32\% fewer pages than baselines, building a knowledge graph of 664 entities and 542 type-consistent relations. Domain-adapted few-shot extraction with a curated glossary and type-constraint definitions eliminates type violations entirely, demonstrating that lightweight domain knowledge integration can substantially improve LLM-based information extraction. The W$\to$K$\to$W approach is general and can be applied to entity discovery in any domain where web sources contain structured but fragmented information.

\subsection{Future Work}

We identify three directions for future research:
\begin{itemize}
    \item \textbf{Active learning for extraction:} Using the KG's own structure to identify high-uncertainty extraction results and selectively request human annotation, reducing LLM error propagation.
    \item \textbf{Temporal dynamics:} Extending the pipeline to track supplier entry and exit over time, enabling supply-chain monitoring rather than one-time mapping.
    \item \textbf{Integration with graph neural networks:} Using GNN-based methods~\cite{wu2021gnn} to learn richer KG representations for gap detection and link prediction, potentially replacing DistMult with more expressive models such as CompGCN or R-GCN. Neural pruning techniques~\cite{ding2025neural} may further improve model efficiency in resource-constrained settings.
\end{itemize}



\begin{thebibliography}{99}

\bibitem{whitehouse_eo14017}
Executive Office of the President. 2021. Executive Order 14017: America's Supply Chains. \emph{Federal Register} 86, 38, 11849--11854. Retrieved from \url{https://www.federalregister.gov/d/2021-04280}

\bibitem{chips_act_2022}
U.S. Congress. 2022. CHIPS and Science Act. Public Law 117-167. Retrieved from \url{https://www.congress.gov/bill/117th-congress/house-bill/4346}

\bibitem{choi2020supply}
Thomas Y. Choi, Dale Rogers, and Bindiya Vakil. 2020. Coronavirus is a wake-up call for supply chain management. \emph{Harvard Business Review} (March 2020). Retrieved from \url{https://hbr.org/2020/03/coronavirus-is-a-wake-up-call-for-supply-chain-management}

\bibitem{chakrabarti1999focused}
Soumen Chakrabarti, Martin van den Berg, and Byron Dom. 1999. Focused crawling: A new approach to topic-specific web resource discovery. \emph{Computer Networks} 31, 11--16, 1623--1640. \url{https://doi.org/10.1016/S1389-1286(99)00052-3}

\bibitem{brintrup2023surveillance}
Alexandra Brintrup, Edward Elson Kosasih, Philip Schaffer, Guanyu Zheng, G\"{u}ne\c{s} Demirel, and Bart L. MacCarthy. 2024. Digital supply chain surveillance using artificial intelligence: Definitions, opportunities and risks. \emph{International Journal of Production Research} 62, 13, 4674--4695. \url{https://doi.org/10.1080/00207543.2023.2270719}

\bibitem{hogan2021knowledge}
Aidan Hogan, Eva Blomqvist, Michael Cochez, Claudia d'Amato, Gerard de Melo, Claudio Gutierrez, Sabrina Kirrane, Jos\'{e} Emilio Labra Gayo, Roberto Navigli, Sebastian Neumaier, Axel-Cyrille Ngonga Ngomo, Axel Polleres, Sabbir M. Rashid, Anisa Rula, Lukas Schmelzeisen, Juan Sequeda, Steffen Staab, and Antoine Zimmermann. 2021. Knowledge graphs. \emph{ACM Computing Surveys} 54, 4, Article 71, 1--37. \url{https://doi.org/10.1145/3447772}

\bibitem{ji2022survey}
Shaoxiong Ji, Shirui Pan, Erik Cambria, Pekka Marttinen, and Philip S. Yu. 2022. A survey on knowledge graphs: Representation, acquisition, and applications. \emph{IEEE Transactions on Neural Networks and Learning Systems} 33, 2, 494--514. \url{https://doi.org/10.1109/TNNLS.2021.3070843}

\bibitem{wei2023zeroshot}
Xiang Wei, Xingyu Cui, Ning Cheng, Xiaobin Wang, Xin Zhang, Shen Huang, Pengjun Xie, Jinan Xu, Yufeng Chen, Meishan Zhang, Yong Jiang, and Wenjuan Han. 2023. Zero-shot information extraction via chatting with ChatGPT. arXiv:2302.10205. Retrieved from \url{https://arxiv.org/abs/2302.10205}

\bibitem{wadhwa2023revisiting}
Somin Wadhwa, Silvio Amir, and Byron Wallace. 2023. Revisiting relation extraction in the era of large language models. In \emph{Proceedings of the 61st Annual Meeting of the Association for Computational Linguistics (ACL~'23)}. Association for Computational Linguistics, Toronto, Canada, 15566--15589. \url{https://doi.org/10.18653/v1/2023.acl-long.868}

\bibitem{pan2024unifying}
Shirui Pan, Linhao Luo, Yufei Wang, Chen Chen, Jiapu Wang, and Xindong Wu. 2024. Unifying large language models and knowledge graphs: A roadmap. \emph{IEEE Transactions on Knowledge and Data Engineering} 36, 7, 3580--3599. \url{https://doi.org/10.1109/TKDE.2024.3352100}

\bibitem{zhu2024llms}
Yuqi Zhu, Xiaohan Wang, Jing Chen, Shuofei Qiao, Yixin Ou, Yunzhi Yao, Shumin Deng, Huajun Chen, and Ningyu Zhang. 2024. LLMs for knowledge graph construction and reasoning: Recent capabilities and future opportunities. \emph{World Wide Web} 27, Article 58. \url{https://doi.org/10.1007/s11280-024-01297-w}

\bibitem{bordes2013translating}
Antoine Bordes, Nicolas Usunier, Alberto Garcia-Dur\'{a}n, Jason Weston, and Oksana Yakhnenko. 2013. Translating embeddings for modeling multi-relational data. In \emph{Advances in Neural Information Processing Systems (NeurIPS~'13)}. 2787--2795. Retrieved from \url{https://papers.nips.cc/paper/2013/hash/1cecc7a77928ca8133fa24680a88d2f9-Abstract.html}

\bibitem{sun2019rotate}
Zhiqing Sun, Zhi-Hong Deng, Jian-Yun Nie, and Jian Tang. 2019. RotatE: Knowledge graph embedding by relational rotation in complex space. In \emph{International Conference on Learning Representations (ICLR~'19)}. Retrieved from \url{https://openreview.net/forum?id=HkgEQnRqYQ}

\bibitem{wichmann2020extracting}
Pascal Wichmann, Alexandra Brintrup, Simon Baker, Philip Woodall, and Duncan McFarlane. 2020. Extracting supply chain maps from news articles using deep neural networks. \emph{International Journal of Production Research} 58, 17, 5320--5336. \url{https://doi.org/10.1080/00207543.2020.1720925}

\bibitem{kosasih2021ml}
Edward Elson Kosasih and Alexandra Brintrup. 2022. A machine learning approach for predicting hidden links in supply chain with graph neural networks. \emph{International Journal of Production Research} 60, 17, 5380--5393. \url{https://doi.org/10.1080/00207543.2021.1956697}

\bibitem{kosasih2022kg}
Edward Elson Kosasih, Federico Margaroli, Simone Gelli, Ajmal Aziz, Nicky Wildgoose, and Alexandra Brintrup. 2023. Towards knowledge graph reasoning for supply chain risk management using graph neural networks. \emph{International Journal of Production Research} 61, 16, 5596--5612. \url{https://doi.org/10.1080/00207543.2022.2100841}

\bibitem{chao1984nonparametric}
Anne Chao. 1984. Nonparametric estimation of the number of classes in a population. \emph{Scandinavian Journal of Statistics} 11, 4, 265--270. Retrieved from \url{https://www.jstor.org/stable/4615964}

\bibitem{chao2005species}
Anne Chao, Robin L. Chazdon, Robert K. Colwell, and Tsung-Jen Shen. 2005. A new statistical approach for assessing similarity of species composition with incidence and abundance data. \emph{Ecology Letters} 8, 2, 148--159. \url{https://doi.org/10.1111/j.1461-0248.2004.00707.x}

\bibitem{bird2019capture}
Sheila M. Bird and Ruth King. 2018. Multiple systems estimation (or capture-recapture estimation) to inform public policy. \emph{Annual Review of Statistics and Its Application} 5, 95--118. \url{https://doi.org/10.1146/annurev-statistics-031017-100641}

\bibitem{bar2006methods}
Ziv Bar-Yossef and Maxim Gurevich. 2008. Random sampling from a search engine's index. \emph{J. ACM} 55, 5, Article 24, 1--74. \url{https://doi.org/10.1145/1411509.1411514}

\bibitem{yang2015embedding}
Bishan Yang, Wen-tau Yih, Xiaodong He, Jianfeng Gao, and Li Deng. 2015. Embedding entities and relations for learning and inference in knowledge bases. In \emph{International Conference on Learning Representations (ICLR~'15)}. Retrieved from \url{https://arxiv.org/abs/1412.6575}

\bibitem{wu2021gnn}
Zonghan Wu, Shirui Pan, Fengwen Chen, Guodong Long, Chengqi Zhang, and Philip S. Yu. 2021. A comprehensive survey on graph neural networks. \emph{IEEE Transactions on Neural Networks and Learning Systems} 32, 1, 4--24. \url{https://doi.org/10.1109/TNNLS.2020.2978386}

\bibitem{qi2024marine}
Yijiashun Qi, Shuzhang Cai, Zunduo Zhao, Jiaming Li, Yanbin Lin, and Zhiqiang Wang. 2024. Benchmarking large language models for image classification of marine mammals. In \emph{2024 IEEE International Conference on Knowledge Graph (ICKG~'24)}. IEEE, 265--274. \url{https://doi.org/10.1109/ICKG63256.2024.00040}

\bibitem{ding2025neural}
Tianqi Ding, Dawei Xiang, Pablo Rivas, and Liang Dong. 2025. Neural pruning for 3D scene reconstruction: Efficient NeRF acceleration. arXiv:2504.00950. Retrieved from \url{https://arxiv.org/abs/2504.00950}

\bibitem{barbaresi2021trafilatura}
Adrien Barbaresi. 2021. Trafilatura: A web scraping library and command-line tool for text discovery and extraction. In \emph{Proceedings of the 59th Annual Meeting of the Association for Computational Linguistics and the 11th International Joint Conference on Natural Language Processing: System Demonstrations (ACL-IJCNLP~'21)}. Association for Computational Linguistics, Online, 122--131. \url{https://doi.org/10.18653/v1/2021.acl-demo.15}

\bibitem{richardson2007beautiful}
Leonard Richardson. 2007. Beautiful Soup: We called him Tortoise because he taught us. Retrieved from \url{https://www.crummy.com/software/BeautifulSoup/}

\bibitem{qi2026sme}
Yijiashun Qi, Hanzhe Guo, and Yijiazhen Qi. 2026. Detecting High-Potential SMEs with Heterogeneous Graph Neural Networks. arXiv:2602.19591. Retrieved from \url{https://arxiv.org/abs/2602.19591}

\bibitem{qi2026llmecon}
Yijiashun Qi, Hanzhe Guo, and Yijiazhen Qi. 2026. Can LLMs Simulate Economic Agents? Testing Production Theory with GPT-Based Firm Behavior. Preprints.

\end{thebibliography}
\end{document}